\documentclass[11pt,a4paper]{article}
\usepackage[hyperref]{acl2021}
\usepackage{times}
\usepackage{latexsym}

\usepackage{amsmath}
\DeclareMathOperator*{\argmax}{arg\,max}

\usepackage{multirow}
\usepackage{graphics}
\usepackage{booktabs}

 \usepackage{graphicx}
\usepackage{microtype}

\aclfinalcopy 
\def\aclpaperid{1444} 

\title{End-to-End NLP Knowledge Graph Construction}

\author{Ishani Mondal\thanks{\hspace{5pt}Work done during internship at IBM Research.}
\\
  Microsoft Research \\
  Bengaluru, India \\
  \texttt{ishani340@gmail.com} \\\And
  Yufang Hou \\
  IBM Research Europe \\
  Dublin, Ireland \\
  \texttt{yhou@ie.ibm.com} \\\And
  Charles Jochim \\
 IBM Research Europe \\
  Dublin, Ireland \\
  \\}

\date{}

\begin{document}
\maketitle
\begin{abstract}
This paper studies the end-to-end construction of an NLP Knowledge Graph (KG) from scientific papers. We focus on extracting four types of relations: \textbf{\emph{evaluatedOn}} between \emph{tasks} and \emph{datasets}, \textbf{\emph{evaluatedBy}} between \emph{tasks} and \emph{evaluation metrics}, as well as \textbf{\emph{coreferent}} and \textbf{\emph{related}} relations between the same type of entities. For instance, ``\emph{F1 score}'' is coreferent with ``\emph{F-measure}''. 
We introduce novel methods for each of these relation types and apply our final framework (\emph{SciNLP-KG}) to 30,000 NLP papers from ACL Anthology to 
build a large-scale KG, which can facilitate automatically constructing scientific leaderboards for the NLP community.  
The results of our experiments indicate that the resulting KG contains high-quality information.
\end{abstract}

\section{Introduction}
As interest in the NLP research community grows, the number of NLP tasks, datasets, and metrics for evaluation also grows, making it increasingly difficult for researchers to keep track of the plethora of new resources.
In order to tackle this problem, recently there have been a few manual efforts to summarize the state-of-the-art on selected subfields of NLP in the form of leaderboards that extract tasks, datasets, metrics and results from papers, such as \emph{NLP-progress}\footnote{https://github.com/sebastianruder/NLP-progress} or \emph{Paperswithcode}\footnote{https://paperswithcode.com}. However, these manual efforts are not sustainable over time for all NLP tasks.

Meanwhile, although there are various studies focusing on extracting entities and relations from scientific literature \cite{augenstein-sogaard-2017-multi,Luan2018,semeval2018, Hou-acl2019,jain-etal-2020-scirex}, there is less work on constructing KGs that contain rich information about tasks, datasets, and metrics. Such a KG would be highly beneficial for the researchers to understand the underlying related literature for a particular task, or to perform comparable experiments.


In this paper, we propose an end-to-end approach to construct a KG from the NLP papers. Our KG contains three types of entities (\emph{\textbf{tasks}}, \emph{\textbf{datasets}}, and \emph{\textbf{metrics}}) and four types of relations connecting them. Figure \ref{fig:framwork} (bottom left) depicts these relations in our large-scale graph built from 30,000 NLP papers from the ACL Anthology. For instance, ``sentiment analysis (task)'' is \emph{\textbf{evaluatedOn}} ``IMDb dataset'', ``opinion analysis (task)'' is \emph{\textbf{evaluatedBy}} ``Precision (metric)'', ``F1-score (metric)'' is \emph{\textbf{coreferent}} with ``F1-measure (metric)'', and ``YELP review dataset'' is \emph{\textbf{related}} to ``IMDb dataset''. 

We develop a framework (\emph{SciNLP-KG}, Section \ref{sec:framework}) to extract these relations based on NLP papers that are tagged for Task (T), Dataset (D), and Metric (M) entities using a TDM entity tagger. Our framework primarily consists of three learning-based models.  First, motivated by \citeauthor{Hou-acl2019}'s work on 
tagging NLP papers with valid TDM triples based on a small manually created gold TDM taxonomy,
we design a hybrid NLI (Natural Language Inference)-based relation extraction model 
to extract \emph{evaluatedOn} and \emph{evaluatedBy} relations.
Our model can extract these two relations at the document level even if the related entities do not appear in the same sentence.  Second, for the \emph{coreferent} relation, we use a mention-pair model to identify the same entities within and across documents. We use a few heuristics to generate training instances, such as that authors often use abbreviations to refer to the common terms (i.e., \emph{NER} -- \emph{Named Entity Recognition}). Third, we propose another model, \emph{term2vec}, which is trained on pseudo-sentences that contain tagged TDM entities from different documents. We use the resulting embeddings to extract the \emph{related} relation between similar type of entities. For instance, ``\emph{semantic role labeling}'' is related to ``\emph{argument identification}'' and ``\emph{GENIA Corpus}'' is related to ``\emph{NCBI Corpus}''.

To evaluate our end-to-end \emph{SciNLP-KG} framework, we manually construct a small-scale 
NLP KG based on our proposed schema (Section \ref{sec:smallNLP-KG}), which contains 85 nodes and 625 links. Experiments show that our system achieves reasonable results for all relation types on this small-scale graph with 
all possible meaningful links manually annotated. 
We further apply our framework on 30,000 NLP papers from ACL Anthology to build a large-scale NLP KG containing 5,374 nodes and 15,762 relations. We evaluate the quality and coverage of the KG by manually assessing random samples and comparing it with \emph{Paperswithcode}. We found that our KG contains high-quality information. 

Overall, the contributions of our work are three-fold. First, we propose and design a new schema that represents knowledge about \emph{tasks} (T), \emph{datasets} (D) and \emph{metrics} (M) in the NLP domain. Second, we develop a novel framework (\emph{SciNLP-KG}) for constructing an NLP KG from the scientific literature in an end-to-end manner. Finally, 
we \emph{automatically} build a large-scale NLP KG that contains high-quality information about the Task-Dataset-Metric (TDM) entities. However, our method is generalized in a way that it could be extended to the domains of computer vision or bioinformatics. Our code and datasets are made publicly available at \url{https://github.com/Ishani-Mondal/SciKG} to fuel further research. 

\section{Related Work}
There is a wealth of research in the NLP community on extracting information from scientific literature. 
Earlier work identified citation contexts and then classified them \cite{teufel-etal-2006-automatic, abu-jbara-etal-2013-purpose, jurgens-etal-2018-measuring}.
Other lines of research include unsupervised approaches for extracting paper 
concepts \citep{gupta-manning-2011-analyzing, Tsai:2013:} and 
keyphrases
\citep{kim-etal-2010-semeval, hasan-ng-2014-automatic}.
Here we are most interested in entity 
and relation extraction in scientific papers
\cite{tateisi2014annotation, augenstein-sogaard-2017-multi, gabor-etal-2018-semeval}.
In SemEval 2017 Task 10, \citet{augenstein-etal-2017-semeval} focused on extracting three types of entities (\emph{Process, Task, and Method}) and two relation types (\emph{hyponym-of} and \emph{synonym-of}).
The dataset from this task 
has been used to explore various neural models for IE on scientific literature \citep{luan-etal-2017-scientific, Ammar2017}.
\citet{Luan2018} 
released 
another dataset which contains annotations for five types of entities and eight types of relations on 500 scientific abstracts.
\citet{Ammar2018} built a literature graph for different domains and they focused on identifying entities and linking them to 
the existing knowledge bases.  
\citet{Hou-acl2019} 
developed a 
system to identify Task-Dataset-Metric triples in NLP papers based on a small set of manually constructed gold TDM triples. 
Recently, \citet{jain-etal-2020-scirex} introduced \emph{SciREX}, a document level IE dataset that encompasses multiple IE tasks, including salient entity identification and document level N-ary relation identification from scientific articles. 
Unlike the above mentioned work, we concentrate on building a TDM KG by leveraging entity recognition and relation extraction within and across different sentences/documents. Our constructed KG can facilitate large-scale NLP leaderboard construction and 
help the researchers to gain insights from NLP literature.
\section{NLP Knowledge Graph Schema}
\label{sec:schema}
In this section, we propose an NLP KG schema which contains three types of entities (i.e., \emph{task}, \emph{dataset}, and \emph{metric}) and four types of relations between them. 
An entity mention is a single or multi-word nominal phrase that represents a task (e.g., \emph{named entity recognition}), dataset (e.g., \emph{IMDB}), or evaluation metric (e.g., \emph{F1-score}) entity. We mainly focus on these three types of entities because they are the core concepts of the NLP community. In addition, they are relatively stable across different papers compared to other types of entities, such as method, result, or experiment.\footnote{
By referring to the stability of the entities, we meant to say that TDM entities are more likely to reflect the collectively shared views in the NLP domain, i.e., researchers will use the same name to refer to the same entity in different papers (e.g., sentiment classification). In contrast, 
method/result/experiment in research papers are non-standardized in nature and very difficult to normalize. For instance, a main method ``LSTM-CNN-sentiment'' in one paper might be referred to in another paper as ``LC-based text classification''.} Similarly, we focus on the following four types of relations between TDM entities that represent the collectively shared view among the NLP researchers: \\
\noindent
\textbf{1. evaluatedOn:} This relation implies that a task $T$ is often evaluated on a dataset $D$ (e.g., \textit{sentiment analysis $\rightarrow$ IMDB}).
\noindent \\
\textbf{2. evaluatedBy:} This relation implies that NLP researchers usually evaluate a task $T$ using metric $M$ (e.g., \textit{Named Entity Recognition $\rightarrow$ F1}).\\
\noindent
\textbf{3. coreferent:} This relation is used to capture the fact that an entity may be referred to differently in the same or different papers, such as \emph{Named Entity Recognition -- NER}, \emph{NCBI dataset -- NCBI corpus}, or \emph{F1 score -- F1 measure}. The \emph{coreferent} relation can help us to canonicalize TDM entities in the constructed KG.\\
\noindent
\textbf{4. related:} Similar to word relatedness, this relation captures all types of associations between the same type of entities.\footnote{In principle, the \emph{related} relation can be applied to different type of TDM entities. Here we only consider it between the same type of entities because the \emph{evaluatedOn} and \emph{evaluatedBy} relations already capture the most prominent relations across TDM entities.}
For instance, ``\emph{semantic role labelling}'' is related to \emph{argument identification} because the latter is a sub-task of the former.
Also ``\emph{GENIA Corpus}'' is related to ``\emph{NCBI Corpus}'' because both datasets are used to develop NER models in the biomedical domain.
In practice, there is not a clear way to define all possible relations between TDM entities, and the \emph{related} relation provides a practical and efficient way to navigate KG. 

\section{Dataset Construction}
We create two new datasets for testing NLP KG construction, both of which are derived from the TDM corpus from \citet{hou2021tdmsci} (see Section \ref{sec:TDM-NLP-Papers}).
The first dataset is a manually constructed small-scale NLP KG according to the schema described in Section \ref{sec:schema}. 
The second dataset is constructed for training a model to extract \emph{evaluatedOn} and \emph{evaluatedBy} relations. 

\subsection{TDM Tagged Corpus}
\label{sec:TDM-NLP-Papers}
Our target entities are abstract objects of type \textbf{Task, Metric, and Dataset (TDM)} which are specific instantiations of the entities in document. 
We use the recently released state-of-the-art TDM tagger \cite{hou2021tdmsci}, trained on the Flair framework \cite{akbik-etal-2019-flair} based on the cased BERT-base embeddings \cite{BERT}, to obtain the mentions of Task, Dataset and Metrics.  This tagger is trained on a corpus of 1,500 sentences taken from the full text of NLP papers, which have been annotated by domain experts for TDM concepts. Finally, it has been applied on 30,000 NLP papers from the ACL Anthology. We refer to the resulting dataset as \emph{TDM-NLP-Papers} for the rest of the paper and it will be used as input for our proposed framework to construct an NLP KG.



\begin{figure*}[t]
\centering
    \includegraphics[width=0.98\textwidth, height=7cm]{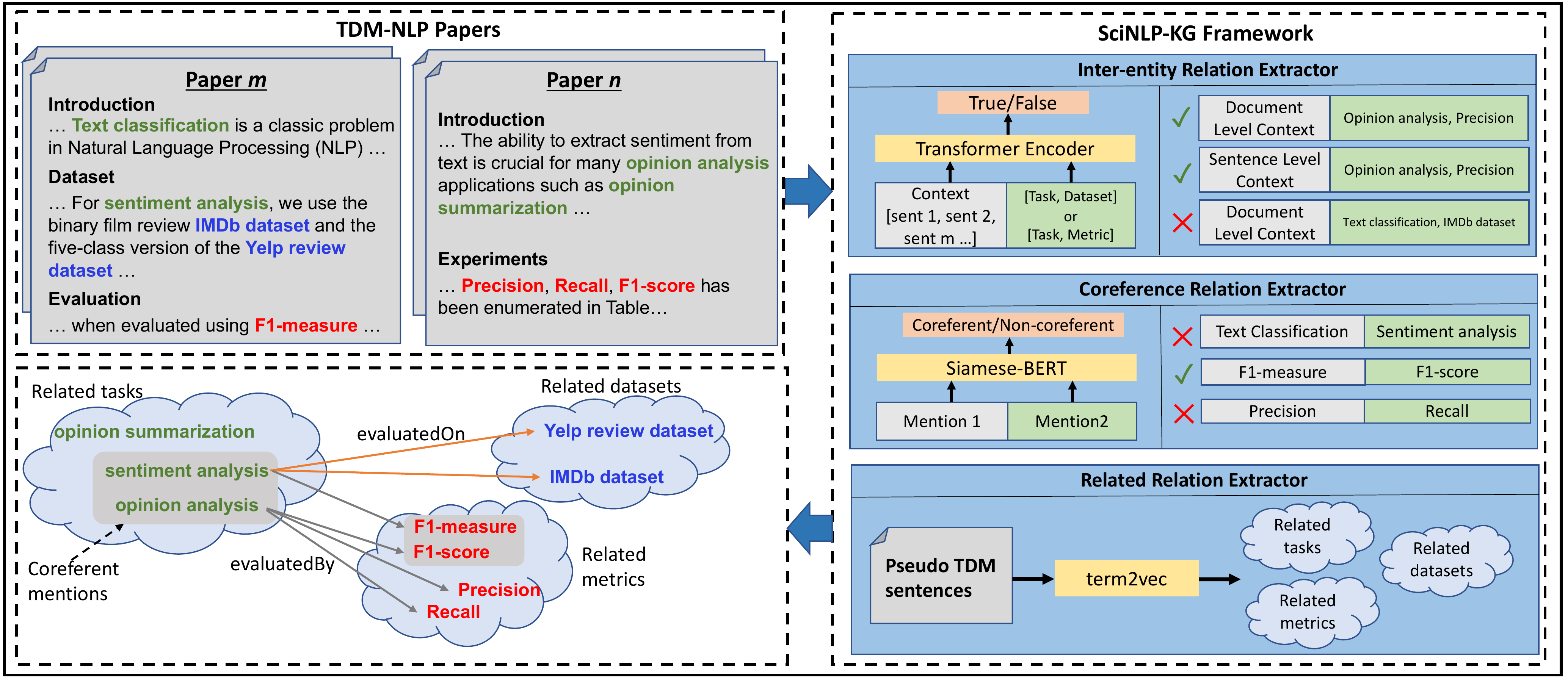} 
    \caption{Overview of the SciNLP-KG Framework}
     \label{fig:framwork}
\end{figure*}

\begin{table}[t]
\small
\centering
\begin{tabular}{lclcc}
\toprule
 \multicolumn{2}{c}{\textbf{Nodes}} &  \multicolumn{2}{c}{\textbf{Relations}}\\
\cmidrule(r){1-2}
\cmidrule(r){3-5}
 \textbf{Type} & \textbf{Count} & \textbf{Type} & \textbf{Count} & \textbf{$\kappa$}\\
\midrule
 Task & 43 & evaluatedOn & 81 & 0.78\\
 Dataset & 23 & evaluatedBy & 249 & 0.73\\
 Metric & 19 & coreferent & 38 & 0.84\\
  &  & related & 257 &  0.94\\
 \midrule
 \textbf{Total} & 85 & \textbf{Total} & 625 \\ 
\bottomrule
\end{tabular}
\caption{Statistics of \emph{smallNLP-KG}. $\kappa$ indicates 
inter-annotator agreement for each relation type.}
\label{tab:stat_small_KG}
\end{table}


\subsection{smallNLP-KG}
\label{sec:smallNLP-KG}
In order to evaluate our approach to construct an NLP KG more efficiently during model development, we have conducted an annotation study to build a small-scale gold NLP KG following the schema given in Section \ref{sec:schema}.

Specifically, the first author sampled a small amount of TDM entities from the tagged dataset \emph{TDM-NLP-Papers}. The chosen entities contain well-established tasks (e.g., \emph{dependency parsing} or \emph{named entity recognition}) and their corresponding datasets as well as evaluation metrics. In order to better facilitate the type of \emph{coreferent} edge in the KG, entities that are abbreviated or have a small edit distance to the existing entities are also added to the list (e.g., \emph{NER}). 

Two domain experts then independently annotate all possible relations described in Section \ref{sec:schema} between any two entities. If necessary, they read the corresponding literature to make a decision. Finally, the annotators reconcile the differences in their annotations and produce the final canonical annotation. The final gold graph contains 85 entities (nodes) and 625 relation instances. Table \ref{tab:stat_small_KG} lists statistics for each entity type and each relation type. We also calculate the inter-annotator agreement per relation type using Cohen's $\kappa$ (see $\kappa$ column in Table \ref{tab:stat_small_KG}). Overall, we achieve high inter-annotator agreement for all the relation types.

Apart from these relations,  we also explored other relations between similar types of entities, such as \emph{hypernym} \cite{hearst-1992-automatic} and \emph{part-of} relations. 
In a pilot annotation study, we found that finer-grained relation types between TDM entities are more difficult to annotate, e.g., the relation between “semantic role labelling” and “argument identification”.
Ultimately, in order to help users to navigate through the built KG, we decide to use the “related” relation to capture all types of associated relations between the same type of entities (e.g., when a user clicks “coreference resolution”, the system can recommend related tasks). 

\subsection{TD/TM Training Dataset}
\label{sec:TD-TM-Rel}
To facilitate the extraction of the \emph{evaluatedBy} and \emph{evaluatedOn} relations, we create a corpus (\emph{TD-TM-Rel}) containing 600 sentences randomly chosen from the tagged \emph{TDM-NLP-Papers} corpus. Each sentence has at least two different types of tagged TDM entities. Two domain experts then annotate the valid \emph{evaluatedBy} and \emph{evaluatedOn} relations for each sentence. The inter-annotator agreements on the  \emph{evaluatedOn} and \emph{evaluatedBy} relations are 0.96 and 0.91 (Cohen's $\kappa$), respectively. Below is an example of TD/TM entities appearing in the same sentence but 
not expressing 
\emph{evalutedOn/evaluatedBy} relations: \emph{“As a testbed for this task, we introduce the \textbf{SentiHood} dataset extracted from a \textbf{question answering} platform where urban neighbourhoods are discussed by users”}. This sentence does not express a valid \emph{evalutedOn} relation between the “question answering” task and the “SentiHood” dataset.


\section{SciNLP-KG Framework}
\label{sec:framework}
In this section, we describe our proposed framework, \emph{SciNLP-KG}, to construct an NLP KG from unstructured text as shown in Figure \ref{fig:framwork}. 
Our framework consists of three models to extract four types of relations as described in Section \ref{sec:schema}. Instead of proposing an end-to-end model (which could suffer from component-wise propagation errors \citep{wadden2019entity,jain-etal-2020-scirex}), we incrementally develop separate components based on the properties of the target relations and the availability of training datasets, and finally aggregate results of each component to create the final KG. 

\subsection{Inter-Entity Relation Extractor}
\label{sec:td/tm-relation-extractor}
The \emph{evaluatedOn} and \emph{evaluatedBy} relations between task and dataset/metric entities depend on the document-level context. It is often the case that the entities involved in these two relations do not occur in the same sentence. On the other hand, a document containing a task mention $t$ and a dataset mention $d$ does not necessarily imply that there exists a positive \emph{evaluatedOn} relation between them. \\
\noindent





\noindent
\textbf{Sentence-Level NLI Model (S-NLI).}
  We train a S-NLI model based on our corpus \emph{TD-TM-Rel} (Section \ref{sec:TD-TM-Rel}). Here \emph{context} is a sentence, and \emph{TD/TM hypothesis} is one of the combinations of TD/TM entities tagged in the sentence. During training, \emph{[CLS]context[SEP]TD/TM hypothesis[SEP]} is given as input to the BERT-based model, followed by sigmoid activation on [CLS] token to predict if the context expresses a valid 
  \emph{evaluatedOn}/\emph{evaluatedBy} relation or not. The model is trained using binary cross-entropy loss. Note that among all combinations, only the ones that correspond to the annotated \emph{evaluatedOn/evaluatedBy} relations are valid TD/TM hypotheses. 
    At inference time, 
  given a task $t$ and a dataset $d$ (or metric $m$), we first extract all sentences from the tagged corpus, \emph{TDM-NLP-Papers}, which contain both $t$ and $d$ (or $m$). 
  We then try to predict whether an \emph{evaluatedOn} (or \emph{evaluatedBy}) relation can be inferred from any of these sentences using the trained model. 


\noindent
\textbf{Document-Level NLI model (D-NLI).} Instead of looking only at the sentence level, we also adopt an approach in which we form the \emph{TD/TM hypothesis} space at the document level.
For example, if a tagged NLP paper has $n$ task entities ($t_1$, $t_2$,.... $t_n$) and $m$  dataset entities ($d_1$, $d_2$,.... $d_m$), we generate $n*m$ combinations of TD tuples as our testing TD hypotheses. For each testing TD hypothesis, we construct \emph{context} by concatenating sentences that contain at least one corresponding entity from the tagged NLP paper. Different from the S-NLI model, the D-NLI model can capture the \emph{evaluatedOn/evaluatedBy} 
relations across sentence boundaries. We use the recently
released \emph{SciREX} corpus \cite{jain-etal-2020-scirex} to train our D-NLI model. \emph{SciREX} contains document-level annotations of $<$ \emph{Task, Dataset, Metric, Method}$>$ tuples as well as the individual entities for 438 NLP papers. Most binary relations such as \emph{Task--Dataset} (\emph{evaluatedOn}) and \emph{Task--Metric} (\emph{evaluatedBy}) occur across sentences. Specifically, for each valid
TD/TM relation annotated at the document level, we treat them as a valid TD/TM hypothesis.  
We form the \emph{context} by concatenating all sentences from the paper that contain the annotations for at least one element in this relation. 
Similarly, we construct negative training instances using other TD/TM combinations by assuming that 4-ary tuple annotations 
in \emph{SciREX} contain all valid TD/TM pairs for a specific paper. \\

\noindent
\textbf{Hybrid NLI Model (H-NLI).}
Although the D-NLI model can capture \emph{evaluatedOn/evaluatedBy} relations across sentences, we assume that the S-NLI model is more accurate because it is easier for the model to learn patterns from shorter contexts. To combine the strengths of both models, we propose a hybrid NLI model. More specifically, given a task $t$ and dataset $d$ (or metric $m$) and the corresponding NLP papers containing these entities, we first apply our S-NLI model to all sentences containing both entities to decide whether an \emph{evaluatedOn} (or \emph{evalautedBy}) relation is held between them. If such a sentence does not exist, we use D-NLI model to make the final prediction. 
By doing so, we cascade the outputs from both S-NLI and D-NLI models in a sieve-based fashion, with the higher priority sieve being the S-NLI model. Thus, it combines the strength of capturing higher precision with S-NLI and better recall with D-NLI.

\subsection{Coreferent Relation Extractor}
\label{sec:coref-relation-extractor}
Unlike coreference resolution on news corpora, which mainly depends on context, we notice that in our scenario, researchers often use abbreviations to refer to common terms (e.g., \emph{NER--Named Entity Recognition}). Sometimes different researchers use slightly different variations to refer to the same entity (e.g., \emph{F1 score--F1}). Motivated by these observations that the surface forms play a pivotal role for TDM coreferent relation extraction, we design a mention-pair model to capture the coreferent relations between the same type of entities.


We generate positive training instances for our mention-pair model using a few heuristics. Specifically, we apply a regular pattern to check whether a tagged entity is followed by its abbreviation in brackets in the tagged NLP papers, such as ``\emph{Named entity recognition (NER)}''. We further extract pairs of entities which have a small edit distance, such as ``\emph{F1 score -- F1}''. Finally, we generate the same number of negative training instances by randomly pairing entities of the same type, which do not meet the criteria of the above heuristics.

We fine-tune our mention-pair model on a \emph{BERT-Siamese Network} \cite{reimers-gurevych-2019-sentence}. We use mean pooling over the output of two [CLS] tokens and use the Euclidean distance function in the penultimate layer, which is followed by a fully connected softmax layer with two labels (\emph{coreferent} and \emph{non-coreferent}). 
We form mention-pairs from within and across documents, so this component can be seen as a cross-document 
coreference resolution 
module.

\subsection{Related Relation Extractor}
\label{sec:related-relation-extractor}
We observe that in our \emph{TDM-NLP-Papers} corpus, the co-occurrence patterns of TDM entities in different contexts (documents) encode rich semantic information about each individual entity. Motivated by \emph{word2vec} \cite{word2vec_Mikolov}, we propose an unsupervised \emph{term2vec} model to capture related relations between TDM entities.

We hypothesize that the TDM entities in a single document are somehow related, even if they do not occur in the same sentence.
We then generate a pseudo-sentence for each tagged paper. 
A pseudo-sentence per paper treats the whole paper as the context which contains special “word”--TDM entities. 
Two such examples are: \emph{“sent0: sentiment analysis, aspect-based sentiment analysis, semeval 2014 task 4 laptop, sentihood, text classification”}, and \emph{“sent1: sentiment classification, semeval 2014 task 4 laptop, sentihood”}. After applying \emph{term2vec} on these pseudo sentences, terms with similar contexts will be close to each other in the embedding space, which can help us to identify related relations between TDM entities. 

Specifically, for each paper in \emph{TDM-NLP-Papers}, we concatenate all tagged entities to form a pseudo sentence. We treat each entity (term) as a single word in this sentence and generate vector representations for each term using the Skip-gram model, which preserves the type of each term.

For a term $j$, the algorithm models the neighborhood of this term as shown in equation (1):

\begin{equation}
   \argmax_\theta \sum_{j \in V} \sum_{i \in K_V} \sum_{c_i \in N_i(j)}\ \log p(c_i|j;\theta) 
\end{equation}

\noindent
Here, $K_V$ denotes the type of nodes. $N_i(j)$ denotes the neighborhood of the term $j$ with respect to the $i^{th}$ type of terms and $p(c_i|j; \theta)$ represents the conditional probability of having a context term $c$ of type $i$ given the term $j$. The objective of this algorithm is to predict the neighborhood of a given term of a particular type $i$ to other terms of similar type. After obtaining the $d$-sized embeddings for each term, we use the unsupervised $K$-means clustering algorithm to determine the clusters for each term. These 
clusters, thus generated, are among the same type of entities and encode the \emph{related} relations among these entities.

\section{Experimental Setup on smallNLP-KG}

 We first evaluate our framework on \emph{smallNLP-KG}.
 We report precision, recall, and F-score fore each relation type. Note that for evaluating \emph{coreferent} and \emph{related} clusters, we consider all pairs of entities in a cluster to be linked.
 \paragraph{Dev/test split.}
 The nodes and relations in \emph{smallNLP-KG} are carefully divided in 10-90\% dev-test split. To avoid any leakage problem as observed in KG completion tasks \cite{sun-etal-2020-evaluation,pezeshkpour2020revisiting}, we exclude training instances from the NLI models and the mention-pair model that involve any of the testing entities. 
More specifically, first, we went through the whole \emph{smallNLP-KG} to make sure any TDM entities (including their coreferent mentions) in the test data do not appear in the dev set. Second, we exclude all these entities (including their coreferent mentions) from the training data for each relation extraction component.
 For the rule-based and unsupervised baselines of \emph{coreferent} and \emph{related} relations (Section \ref{sec:corefrelation} and Section \ref{sec:relatedRelation}), we tune our parameters on the dev set and report results for all relations on the test set.

\paragraph{Implementation details.}
Given all testing entities from \emph{smallNLP-KG} as the input, we use our framework presented in the previous section to build the KG. 

For the hybrid NLI model to extract inter-entity relations (Section \ref{sec:td/tm-relation-extractor}), we fine-tune all NLI models for 3 epochs with a learning rate of $5e-5$ and a batch size of 32. We also carry out experiments with different pre-trained contextual embeddings: \emph{BERT-Base}, \emph{BERT-Large}, \emph{RoBERTa-Base}, as well as \emph{scibert-scivocab-cased} \cite{beltagy-etal-2019-scibert} from the PyTorch-Transformers library. During the inference stage, let there be $n_t$ tasks, $n_d$ datasets and $n_m$ metrics in the \emph{smallNLP-KG} test corpus, so we can have total ($n_t$ $\times$ $n_d$) and ($n_t$ $\times$ $n_m$) combinations of possible \emph{evaluatedOn} and \emph{evaluatedBy} relations respectively. We apply our trained hybrid NLI model to test all combinations.

For the \emph{term2vec} model to extract related relations (Section \ref{sec:related-relation-extractor}), we  
use the Skip-gram word2vec implementation from Gensim 
with a window size of 5, min count of 1, and an embedding dimension of 100. We run the model on all pseudo sentences generated from the whole \emph{TDM-NLP-Papers} corpus. After obtaining term embeddings, we use $K$-Means clustering algorithm from Scikit-learn 
to generate clusters among the same type of entities based on nodes from \emph{smallNLP-KG}. We set $K$ equal to number of gold clusters in \emph{smallNLP-KG} for entity type.

\begin{table}[t]
\centering
\small
\begin{tabular}{l ccc ccc}
\toprule
 & \multicolumn{3}{c}{\textbf{evaluatedOn}} & \multicolumn{3}{c}{\textbf{evaluatedBy}}\\
\cmidrule(r){2-4}
\cmidrule(r){5-7}
\textbf{Methods} & $P$ & $R$ & $F1$ & $P$ & $R$ & $F1$\\
\midrule
\textbf{Sent-Level} &
 &  &  &
 &  &  \\
-BERT-S &
0.77 & 0.23 & 0.35 & 
0.68 & 0.25 & 0.36 \\
-BERT-L &
0.80 & 0.23 & 0.36 & 
\textbf{0.71} & 0.27 &  0.39\\
-SciBERT &
0.81  & 0.22 & 0.35 & 
0.70 & 0.22 & 0.33\\
-RoBERTa &
\textbf{0.83} & 0.24 & 0.37 &
0.70 & 0.27 & 0.39 \\
\bottomrule
\textbf{Doc-Level} &
 &  &  &
&  &  \\
-BERT-S &
0.51  & 0.80 & 0.62 & 
0.44 & 0.68 & 0.53 \\
-BERT-L &
0.52  & 0.80 & 0.63 & 
0.45 & 0.71 & 0.55\\
-SciBERT &
0.57  & 0.84 & 0.68 & 
0.43 & \textbf{0.73} & 0.54\\
-RoBERTa &
0.62 & \textbf{0.86} & 0.73 &
0.45 & 0.72 & 0.55 \\
\bottomrule
\textbf{Hybrid} &
 &  &  &
 &  &  \\
-BERT-S &
0.60  & 0.78 & 0.68 & 
0.50 & 0.64 & 0.56 \\
-BERT-L &
0.63  & 0.80 & 0.70 & 
0.51 & 0.65 & 0.57\\
-SciBERT &
0.64  & 0.81 & 0.72 & 
0.48 & 0.67 & 0.56\\
-RoBERTa &
0.67 & 0.82 & \textbf{0.74} &
0.59 & 0.69 & \textbf{0.64} \\
\bottomrule
\end{tabular}
\caption{Ablation study of NLI Models for extracting  \textit{evaluatedOn}, \textit{evaluatedBy} relations on \emph{smallNLP-KG}.
\label{tab:results_NLI_ablation}
}
\end{table}

\section{Results on smallNLP-KG}

\subsection{Comparison With the Existing Baselines}
\label{sec:baselines}
We compare our \emph{SciNLP-KG} framework against a few baselines from previous work:
\paragraph{1. E2E Rel-Gold \cite{miwa-bansal-2016-end}.} 
The TDM entities present in \emph{TDM-NLP-Papers} are being fed as input to this LSTM-based model, which uses word sequences and dependency tree structures to predict relations. We train the model using our TD/TM training dataset and perform intra-sentence \emph{evaluatedOn/EvaluatedBy} relation extraction using this model. 
\paragraph{2. E2E Rel
\cite{miwa-bansal-2016-end}.} This is another variant of the previous model, which predicts the TD and TM relations in an end-to-end fashion without using the TDM entities tagged in \emph{TDM-NLP-Papers}. We train this model to tag the TDM entities and predict the relations in an end-to-end fashion instead of providing the gold mentions as input to the relation extractor.
\paragraph{3. DocTAET \cite{Hou-acl2019}.}
This model is analogous to our D-NLI model which works in document-level. We use the NLI model proposed in this paper for predicting TD and TM binary relations individually for comparable experimental evaluation, without considering them as a 3-ary relation between $<$Task, Dataset, Metric$>$). 
\paragraph{4. E2E Coref \cite{lee2017endtoend}.}
For coreference resolution, we consider the nouns as we consider only three types of nominal entities. We make use of the end-to-end BiLSTM-CRF based architecture using ELMO embeddings optimized using the conditional likelihood of predicting the correct antecedent given a mention. In our experiments, we used the noun phrase coreference resolution only and discarded the pronoun-based coreference resolution component.
\paragraph{5. SciIE-Rel \cite{Luan2018}.}
This is a multi-task model to extract coreferent and inter-entity relations from the scientific abstracts. We retrain this model on our training datasets and use it to predict \emph{evaluatedOn/evaluatedBy} relations based on the predicted Task, Dataset, and Metric mentions from the \emph{NLP-TDMS-Papers} corpus. 
\paragraph{6. SciIE-Coref \cite{Luan2018}.} This is a component of the above mentioned \emph{SciIE-Rel} system. 
We use this component to extract the coreferent clusters of the same types of entities from the predicted Task, Dataset and Metric mentions of the \emph{NLP-TDMS-Papers} corpus.  


\subsection{Results on Inter-Entity Relations}
Table \ref{tab:results_NLI} shows the results of our hybrid NLI model together with the S-NLI and D-NLI models to extract \emph{evaluatedOn} and \emph{evaluatedBy} relations on the  \emph{smallNLP-KG} test set. The ablation analysis using different embeddings is tabulated in Table \ref{tab:results_NLI_ablation}.
It seems
that the S-NLI model suffers from lower recalls, with 0.24 on the \emph{evaluatedOn} relation and 0.27 on the \emph{evaluatedBy} relation. Exploring the D-NLI context, we observe a drop of 0.21 and 0.25 in terms of precision  on \emph{evaluatedOn} and \emph{evaluatedBy} relations respectively, which is mainly attributed to the relatively longer \emph{context} 
in the NLI model. There is a significant rise in recall for both relations, with improvements of 0.86 and 0.72 on the \emph{evaluatedOn} and \emph{evaluatedBy} relations, respectively. Finally, our H-NLI model combines the strengths of both S-NLI and D-NLI models, with decent improvement in terms of precision compared to the D-NLI architecture, while preserving better recall than S-NLI model.

\begin{table}[t]
\small
\centering
\begin{tabular}{l ccc ccc}
\toprule
 & \multicolumn{3}{c}{\textbf{evaluatedOn}} & \multicolumn{3}{c}{\textbf{evaluatedBy}}\\
\cmidrule(r){2-4}
\cmidrule(r){5-7}
\textbf{Methods} & $P$ & $R$ & $F1$ & $P$ & $R$ & $F1$\\
\midrule
Baselines &
 &  &  &
 &  &  \\
E2E Rel &
 0.58 & 0.62 & 0.60 &
0.45 & 0.54 & 0.49  \\

E2E Rel-Gold &
 0.63 & 0.69 & 0.65 &
0.51 & 0.58 & 0.54  \\
SciIE Rel &
 0.68 & 0.78 & 0.72 &
0.51 & 0.62 & 0.56  \\
DOCTAET & 0.62 & \textbf{0.86} & 0.73 &
0.45 & \textbf{0.72} & 0.55 \\
\midrule
S-NLI &\textbf{0.83} & 0.24 & 0.37 &
\textbf{0.70} & 0.27 & 0.39 \\
D-NLI &
0.62 & \textbf{0.86} & 0.73 &
0.45 & \textbf{0.72} & 0.55 \\
H-NLI &
 0.67 & 0.82 & \textbf{0.74} &
0.59 & 0.69 & \textbf{0.64}  \\
\bottomrule
\end{tabular}
\caption{
Results of \emph{SciNLP-KG} for extracting  \textit{evaluatedOn} and \textit{evaluatedBy} relations and comparison with the best-performing variant of existing baselines.
\label{tab:results_NLI}
}
\end{table}

\begin{table}[t]
\centering
\small
\begin{tabular}{l ccc ccc}
\toprule
 & \multicolumn{3}{c}{\textbf{coreferent}} & \multicolumn{3}{c}{\textbf{related}}\\
\cmidrule(r){2-4}
\cmidrule(r){5-7}
\textbf{Methods} & $P$ & $R$ & $F1$ & $P$ & $R$ & $F1$\\
\midrule
Baselines &
 &  &  &
 &  &  \\
Rule-based &

0.43 & 0.58 & 0.49 &
- & - & - \\
E2E Coref &
0.61 & 0.67 & 0.64 &
- & - & - \\
SciIE Coref &

0.65 & 0.70 & 0.69 &
 - & - & - \\
PMI &

- & - & - &
 0.61 & 0.61 & 0.61 \\
\midrule
SciNLP-KG &\textbf{0.77} & \textbf{0.79} & \textbf{0.77} &
\textbf{0.67} & \textbf{0.68} & \textbf{0.67}  \\
\bottomrule
\end{tabular}
\caption{Results of \emph{SciNLP-KG} for \textit{coreferent} and \emph{related} relation extraction compared to baselines.
\label{tab:corefresults}
}
\end{table}



\subsection{Results on  \emph{coreferent} Relations}
\label{sec:corefrelation}
We compare our mention-pair model (Section \ref{sec:coref-relation-extractor}) to two coreference resolution systems (\emph{E2E Coref} and \emph{SciIE Coref} in Section \ref{sec:baselines}). We also implement
a heuristic baseline that uses the Jaccard similarity of two entity mentions and predicts those with score greater than 0.75 as coreferent.

Table \ref{tab:corefresults} (left) lists the results of our mention-pair model for \emph{coreferent} relation extraction on the \emph{smallNLP-KG} test set, with an overall Macro F1 of 0.77. 
Overall, we found that our system 
outperforms the 
other three baselines
slightly. 
During qualitative analysis, we observe that our learning-based model eliminates some false positive links proposed by the rule-based approach, such as $<$\emph{Rouge-1} \textbf{coreferent} \emph{Rough-2}$>$ and also captures more true positive links such as $<$\textit{sentiment mining} \textbf{coreferent} \textit{sentiment analysis}$>$. 

\subsection{Results on \emph{related} Relations}
\label{sec:relatedRelation}
We compare our \emph{term2vec} model (Section \ref{sec:related-relation-extractor}) to a baseline model which use \emph{pointwise mutual information} (PMI) \cite{church-hanks-1990-word} to measure the association score between two TDM entities of the same type. 

Table \ref{tab:corefresults} (right) lists the results of our \emph{term2vec} model for extracting \emph{related} relations on \emph{smallNLP-KG}. 
We found that our \emph{term2vec} model 
outperforms the PMI-based baseline on \emph{task} and \emph{metric} entities by a good margin. 
We also observed that there is a slight 
improvement on the dataset clusters. This happens because the datasets inside the same documents are related most of the time, whereas the same is not always true for tasks and metrics, in which the context window of their mentions plays an important role to capture whether they are \emph{related}. During qualitative analysis, we observe some false positives generated from the PMI-based baseline such as $<$\textit{f1-score} \textbf{related} \textit{rouge}$>$. Interestingly, we find that our \emph{term2vec} model captures both coreferent relations as well as other related relations between the same type of entities, such as \emph{hypernym} relations (e.g,  
$<$\textit{parsing} \textbf{related} \textit{dependency parsing}$>$, $<$\textit{rouge} \textbf{related} \textit{rouge-n}$>$). 


\subsection{Comparison with the Existing Models}
Our final Hybrid-NLI model performs better than the two recent relation extraction systems (\emph{E2E Rel} and \emph{SciIE Rel}), similarly our cross-document mention-pair coreference resolver outperforms two related baselines. 
Probing deeper, we observe that existing baselines struggle because: 1) they fail to resolve certain long-range relations due to the end-to-end setup and suffer from error propagation; and 2) they cannot handle inter-document coreference resolution, failing to generate some $<$Task, Dataset$>$ pairs found when canonicalizing entities in cross-document discourse. \citet{Hou-acl2019}'s method suffers from low precision due to the bottleneck of the D-NLI approach (see Table \ref{tab:results_NLI}).
Our hybrid NLI approach achieves better precision while still keeping a reasonable recall. 

It is worth noting that 
in \emph{smallNLP-KG} we annotated completed relations for every pair of TDM mentions.
The annotations already take care of inference-based KG consistency (e.g., T1 \emph{coreferent} T2, T1 \emph{evaluatedOn} D1, T2 \emph{evaluatedOn} D1).
In general, our framework handles each of the relations separately and achieves good performance compared to the baselines which use joint modelling approaches
on the \emph{smallNLP-KG} testing set.


\section{Experiments on LargeNLP-KG}
We apply our trained \emph{SciNLP-KG} framework to the \emph{TDM-NLP-Papers} corpus (Section \ref{sec:TDM-NLP-Papers}) to build a large-scale KG (\emph{largeNLP-KG}). The resulting KG contains 5,374 TDM entities and 15,762 relations (see Table \ref{tab:largestats}). We use both human evaluation as well as automatic evaluation to assess the quality and coverage of \emph{largeNLP-KG}.


\begin{table}[t]
\centering
\small
\begin{tabular}{lclc}
\toprule
 \multicolumn{2}{c}{\textbf{Nodes}} &  \multicolumn{2}{c}{\textbf{Relations}}\\
\cmidrule(r){1-2}
\cmidrule(r){3-4}
 \textbf{Type} & \textbf{Count} & \textbf{Type} & \textbf{Count}\\
\midrule
 Task & 2,441 & evaluatedOn & 4,137 \\
 Dataset & 2,314 & evaluatedBy & 4,019 \\
 Metric & 619 & coreferent & 2,074 \\
  &  & related & 5,532 \\
 \midrule
 \textbf{Total} & 5,374 & \textbf{Total} & 15,762 \\ 
\bottomrule
\end{tabular}
\caption{Overall statistics of \emph{largeNLP-KG}. 
\label{tab:largestats}
}
\end{table}

\paragraph{Human Evaluation.} We randomly sample 100 instances from the final large-scale KG for \emph{evaluatedOn}, \emph{evaluatedBy}, and \emph{coreferent} relations, which were then manually assessed by an NLP expert. 
Note that these relation instances do not appear in the training datasets of our \emph{SciNLP-KG} system. Table \ref{tab:sampled_results} reports the precision for each relation type. It is encouraging to see that our large-scale NLP KG obtains reasonably high precision (0.84, 0.77 and 0.79) on the \emph{evaluatedOn}, \emph{evaluatedBy} and \emph{coreferent} relations, respectively. We found that most  false  positives  are  from the TDM tagger errors. For instance, \emph{Stanford-CoreNLP} is tagged as a \emph{dataset} entity. 

In case of \emph{related} relation clusters, we randomly sample 10 entities from each entity type (i.e., \emph{Task, Dataset, Metric}) and choose the top  20 nearest neighbors of the same type based on the cosine-similarity between entities. An NLP expert assessed the correctness of these 600 relations (200 for each of the T-T, D-D, and M-M related relations) based on their common sense knowledge about NLP. We report $Macro-precision@K$, where $K$= 5, 10, 20 in Table \ref{tab:sampled_results} (last row). In general, for a given entity, our unsupervised \emph{term2vec} model provides reasonable suggestions of the related entities.

\begin{table}[t]
\centering
\small
\begin{tabular}{l c c c c c}
\toprule
 Relations & $\#$ & $Prec$ & $P@5$ & $P@10$ & $P@20$\\
\midrule
\textbf{evaluatedOn} & 100
 & 0.84 & - & - & -\\

\textbf{evaluatedBy} & 100
 & 0.77 & - & - & -\\
\textbf{coreferent} & 100
 & 0.79 & - & - & -\\
 \textbf{related} & 600
 & - & 0.77 & 0.71 & 0.60\\
\bottomrule
\end{tabular}
\caption{Precision of the randomly selected samples from \emph{largeNLP-KG}. $\#$ indicates sample size.
\label{tab:sampled_results}
}
\end{table}

\begin{table}[!t]
\centering
\small
\begin{tabular}{l c c}
\toprule
Method& evaluatedOn & evaluatedBy\\
\midrule
\emph{Relaxed Match} &
0.49 & 0.58 \\
+\emph{coreferent} &
0.56 & 0.63 \\
\bottomrule
\end{tabular}
\caption{\begin{small}Coverage of \emph{largeNLP-KG} compared with \emph{Paperswithcode} on \textit{evaluatedOn/evaluatedBy}  relations.\end{small} 
\label{tab:coverage}
}

\end{table}

\paragraph{Automatic Evaluation.} In order to better understand the coverage of \emph{largeNLP-KG}, we compared \emph{evaluatedOn} and \emph{evaluatedBy} relations from \emph{largeNLP-KG} with the manually constructed NLP leaderboards in \emph{Paperswithcode}.\footnote{We do not evaluate our system on SciREX because we use annotations from SciREX to train our D-NLI model.} 

The recent version of \emph{Paperswithcode} (Aug-2020) contains leaderboard information for 265 NLP tasks and the corresponding 100 datasets. We consider only the NLP-related TDM entities. Each leaderboard is a tuple of four elements (\emph{$<$task, dataset, metric, score$>$}) and it encodes the valid relations between the task and the dataset/metric, which corresponds to our \emph{evaluatedOn} and \emph{evaluatedBy} relations. In total, we obtain 450 TD pairs and 623 TM pairs from \emph{Paperswithcode}. 

We automatically check how many of these pairs are encoded in our large-scale NLP KG. Note that we do not compare TD/TM pairs that appear in the training dataset for our NLI models (Section \ref{sec:td/tm-relation-extractor}).
Specifically, we use an edit-distance matching algorithm to match TD/TM pairs between \emph{largeNLP-KG} and \emph{Paperswithcode} (\emph{Relaxed Match}). We consider the edit-distance of our extracted TD and TM pairs with those in \emph{Paperswithcode} and choose those with normalized edit-distance less than 0.3 as positive instances. For instance, the \emph{Paperswithcode} TM pair ``\emph{sentiment mining}--\emph{F1-score}'' is equivalent to extracted TM pair ``\emph{sentiment analysis}--\emph{F1-scores}''. 

Table \ref{tab:coverage} shows that with \emph{Relaxed Match}, our \emph{largeNLP-KG} contains 49\% and 58\% of TD and TM pairs from \emph{Paperswithcode}, respectively. We further employ coreferent relations to generate more \emph{evaluatedOn} and \emph{evaluatedBy} relations, which helps to achieve a higher coverage of 56\% and 63\% on \emph{Paperswithcode} on TD and TM pairs respectively.  
The mismatch part between our large KG and \emph{Paperswithcode} is mostly due to the fact that \emph{Paperswithcode} contains recent papers while our KG is built on papers from 1974-2019. For instance, our graph contains a task called “textual entailment”, which is not included as a task entity in \emph{Paperswithcode}. 



\section{Conclusions}
In this paper, we propose \emph{SciNLP-KG} framework to build a large-scale NLP KG from NLP papers in an end-to-end manner.
An interesting direction of further research is the diachronic analysis of TDM entities. For instance, for the two tasks-``textual entailment" and ``NLI", it seems that after the emergence of SNLI corpus paper \citep{bowman-etal-2015-large}, the NLP community switched to use the new name to refer to same task. In practice, the automatically-built KG (\emph{largeNLP-KG}) has the potential to assist the researchers to search related papers and develop comparable experiments. In future, we plan to build a web-based visualization tool to enable researchers to explore KG and related papers.

\section*{Acknowledgement}
The authors would gratefully like to thank the anonymous reviewers for their insightful feedback and comments towards improving the paper.

\bibliographystyle{acl_natbib}
\bibliography{anthology,acl2021}

\begin{thebibliography}{32}
\expandafter\ifx\csname natexlab\endcsname\relax\def\natexlab#1{#1}\fi

\bibitem[{Abu-Jbara et~al.(2013)Abu-Jbara, Ezra, and
  Radev}]{abu-jbara-etal-2013-purpose}
Amjad Abu-Jbara, Jefferson Ezra, and Dragomir Radev. 2013.
\newblock \href {https://www.aclweb.org/anthology/N13-1067} {Purpose and
  polarity of citation: Towards {NLP}-based bibliometrics}.
\newblock In \emph{Proceedings of the 2013 Conference of the North {A}merican
  Chapter of the Association for Computational Linguistics: Human Language
  Technologies}, pages 596--606, Atlanta, Georgia. Association for
  Computational Linguistics.

\bibitem[{Akbik et~al.(2019)Akbik, Bergmann, Blythe, Rasul, Schweter, and
  Vollgraf}]{akbik-etal-2019-flair}
Alan Akbik, Tanja Bergmann, Duncan Blythe, Kashif Rasul, Stefan Schweter, and
  Roland Vollgraf. 2019.
\newblock \href {https://doi.org/10.18653/v1/N19-4010} {{FLAIR}: An easy-to-use
  framework for state-of-the-art {NLP}}.
\newblock In \emph{Proceedings of the 2019 Conference of the North {A}merican
  Chapter of the Association for Computational Linguistics (Demonstrations)},
  pages 54--59, Minneapolis, Minnesota. Association for Computational
  Linguistics.

\bibitem[{Ammar et~al.(2018)Ammar, Groeneveld, Bhagavatula, Beltagy, Crawford,
  Downey, Dunkelberger, Elgohary, Feldman, Ha, Kinney, Kohlmeier, Lo, Murray,
  Ooi, Peters, Power, Skjonsberg, Wang, Willhelm, Yuan, van Zuylen, and
  Etzioni}]{Ammar2018}
Waleed Ammar, Dirk Groeneveld, Chandra Bhagavatula, Iz~Beltagy, Miles Crawford,
  Doug Downey, Jason Dunkelberger, Ahmed Elgohary, Sergey Feldman, Vu~Ha,
  Rodney Kinney, Sebastian Kohlmeier, Kyle Lo, Tyler Murray, Hsu-Han Ooi,
  Matthew Peters, Joanna Power, Sam Skjonsberg, Lucy Wang, Chris Willhelm,
  Zheng Yuan, Madeleine van Zuylen, and Oren Etzioni. 2018.
\newblock \href {https://doi.org/10.18653/v1/N18-3011} {Construction of the
  literature graph in semantic scholar}.
\newblock In \emph{Proceedings of the 2018 Conference of the North American
  Chapter of the Association for Computational Linguistics: Human Language
  Technologies, Volume 3 (Industry Papers)}, pages 84--91. Association for
  Computational Linguistics.

\bibitem[{Ammar et~al.(2017)Ammar, Peters, Bhagavatula, and Power}]{Ammar2017}
Waleed Ammar, Matthew Peters, Chandra Bhagavatula, and Russell Power. 2017.
\newblock \href {https://doi.org/10.18653/v1/S17-2097} {The ai2 system at
  semeval-2017 task 10 (scienceie): semi-supervised end-to-end entity and
  relation extraction}.
\newblock In \emph{Proceedings of the 11th International Workshop on Semantic
  Evaluation (SemEval-2017)}, pages 592--596. Association for Computational
  Linguistics.

\bibitem[{Augenstein et~al.(2017)Augenstein, Das, Riedel, Vikraman, and
  McCallum}]{augenstein-etal-2017-semeval}
Isabelle Augenstein, Mrinal Das, Sebastian Riedel, Lakshmi Vikraman, and Andrew
  McCallum. 2017.
\newblock \href {https://doi.org/10.18653/v1/S17-2091} {{S}em{E}val 2017 task
  10: {S}cience{IE} - extracting keyphrases and relations from scientific
  publications}.
\newblock In \emph{Proceedings of the 11th International Workshop on Semantic
  Evaluation ({S}em{E}val-2017)}, pages 546--555, Vancouver, Canada.
  Association for Computational Linguistics.

\bibitem[{Augenstein and S{\o}gaard(2017)}]{augenstein-sogaard-2017-multi}
Isabelle Augenstein and Anders S{\o}gaard. 2017.
\newblock \href {https://doi.org/10.18653/v1/P17-2054} {Multi-task learning of
  keyphrase boundary classification}.
\newblock In \emph{Proceedings of the 55th Annual Meeting of the Association
  for Computational Linguistics (Volume 2: Short Papers)}, pages 341--346,
  Vancouver, Canada. Association for Computational Linguistics.

\bibitem[{Beltagy et~al.(2019)Beltagy, Lo, and
  Cohan}]{beltagy-etal-2019-scibert}
Iz~Beltagy, Kyle Lo, and Arman Cohan. 2019.
\newblock \href {https://doi.org/10.18653/v1/D19-1371} {{S}ci{BERT}: A
  pretrained language model for scientific text}.
\newblock In \emph{Proceedings of the 2019 Conference on Empirical Methods in
  Natural Language Processing and the 9th International Joint Conference on
  Natural Language Processing (EMNLP-IJCNLP)}, pages 3615--3620, Hong Kong,
  China. Association for Computational Linguistics.

\bibitem[{Bowman et~al.(2015)Bowman, Angeli, Potts, and
  Manning}]{bowman-etal-2015-large}
Samuel~R. Bowman, Gabor Angeli, Christopher Potts, and Christopher~D. Manning.
  2015.
\newblock \href {https://doi.org/10.18653/v1/D15-1075} {A large annotated
  corpus for learning natural language inference}.
\newblock In \emph{Proceedings of the 2015 Conference on Empirical Methods in
  Natural Language Processing}, pages 632--642, Lisbon, Portugal. Association
  for Computational Linguistics.

\bibitem[{Church and Hanks(1990)}]{church-hanks-1990-word}
Kenneth~Ward Church and Patrick Hanks. 1990.
\newblock \href {https://www.aclweb.org/anthology/J90-1003} {Word association
  norms, mutual information, and lexicography}.
\newblock \emph{Computational Linguistics}, 16(1):22--29.

\bibitem[{Devlin et~al.(2019)Devlin, Chang, Lee, and Toutanova}]{BERT}
Jacob Devlin, Ming-Wei Chang, Kenton Lee, and Kristina Toutanova. 2019.
\newblock \href {https://doi.org/10.18653/v1/N19-1423} {{BERT}: Pre-training of
  deep bidirectional transformers for language understanding}.
\newblock In \emph{Proceedings of the 2019 Conference of the North {A}merican
  Chapter of the Association for Computational Linguistics: Human Language
  Technologies, Volume 1 (Long and Short Papers)}, pages 4171--4186,
  Minneapolis, Minnesota. Association for Computational Linguistics.

\bibitem[{G{\'{a}}bor et~al.(2018)G{\'{a}}bor, Buscaldi, Schumann, QasemiZadeh,
  Zargayouna, and Charnois}]{semeval2018}
Kata G{\'{a}}bor, Davide Buscaldi, Anne{-}Kathrin Schumann, Behrang
  QasemiZadeh, Ha{\"{\i}}fa Zargayouna, and Thierry Charnois. 2018.
\newblock \href {https://aclanthology.info/papers/S18-1111/s18-1111}
  {Semeval-2018 task 7: Semantic relation extraction and classification in
  scientific papers}.
\newblock In \emph{Proceedings of The 12th International Workshop on Semantic
  Evaluation, SemEval@NAACL-HLT, New Orleans, Louisiana, June 5-6, 2018}, pages
  679--688.

\bibitem[{G{\'a}bor et~al.(2018)G{\'a}bor, Buscaldi, Schumann, QasemiZadeh,
  Zargayouna, and Charnois}]{gabor-etal-2018-semeval}
Kata G{\'a}bor, Davide Buscaldi, Anne-Kathrin Schumann, Behrang QasemiZadeh,
  Ha{\"\i}fa Zargayouna, and Thierry Charnois. 2018.
\newblock \href {https://doi.org/10.18653/v1/S18-1111} {{S}em{E}val-2018 task
  7: Semantic relation extraction and classification in scientific papers}.
\newblock In \emph{Proceedings of The 12th International Workshop on Semantic
  Evaluation}, pages 679--688, New Orleans, Louisiana. Association for
  Computational Linguistics.

\bibitem[{Gupta and Manning(2011)}]{gupta-manning-2011-analyzing}
Sonal Gupta and Christopher Manning. 2011.
\newblock \href {https://www.aclweb.org/anthology/I11-1001} {Analyzing the
  dynamics of research by extracting key aspects of scientific papers}.
\newblock In \emph{Proceedings of 5th International Joint Conference on Natural
  Language Processing}, pages 1--9, Chiang Mai, Thailand. Asian Federation of
  Natural Language Processing.

\bibitem[{Hasan and Ng(2014)}]{hasan-ng-2014-automatic}
Kazi~Saidul Hasan and Vincent Ng. 2014.
\newblock \href {https://doi.org/10.3115/v1/P14-1119} {Automatic keyphrase
  extraction: A survey of the state of the art}.
\newblock In \emph{Proceedings of the 52nd Annual Meeting of the Association
  for Computational Linguistics (Volume 1: Long Papers)}, pages 1262--1273,
  Baltimore, Maryland. Association for Computational Linguistics.

\bibitem[{Hearst(1992)}]{hearst-1992-automatic}
Marti~A. Hearst. 1992.
\newblock \href {https://www.aclweb.org/anthology/C92-2082} {Automatic
  acquisition of hyponyms from large text corpora}.
\newblock In \emph{{COLING} 1992 Volume 2: The 15th {I}nternational
  {C}onference on {C}omputational {L}inguistics}.

\bibitem[{Hou et~al.(2019)Hou, Jochim, Gleize, Bonin, and
  Ganguly}]{Hou-acl2019}
Yufang Hou, Charles Jochim, Martin Gleize, Francesca Bonin, and Debasis
  Ganguly. 2019.
\newblock \href {https://www.aclweb.org/anthology/P19-1513/} {Identification of
  tasks, datasets, evaluation metrics, and numeric scores for scientific
  leaderboards construction}.
\newblock In \emph{Proceedings of the 57th Conference of the Association for
  Computational Linguistics, {ACL} 2019, Florence, Italy, July 28- August 2,
  2019, Volume 1: Long Papers}, pages 5203--5213.

\bibitem[{Hou et~al.(2021)Hou, Jochim, Gleize, Bonin, and
  Ganguly}]{hou2021tdmsci}
Yufang Hou, Charles Jochim, Martin Gleize, Francesca Bonin, and Debasis
  Ganguly. 2021.
\newblock \href {https://www.aclweb.org/anthology/2021.eacl-main.59} {{TDMS}ci:
  A specialized corpus for scientific literature entity tagging of tasks
  datasets and metrics}.
\newblock In \emph{Proceedings of the 16th Conference of the European Chapter
  of the Association for Computational Linguistics: Main Volume}, pages
  707--714, Online. Association for Computational Linguistics.

\bibitem[{Jain et~al.(2020)Jain, van Zuylen, Hajishirzi, and
  Beltagy}]{jain-etal-2020-scirex}
Sarthak Jain, Madeleine van Zuylen, Hannaneh Hajishirzi, and Iz~Beltagy. 2020.
\newblock \href {https://doi.org/10.18653/v1/2020.acl-main.670} {{S}ci{REX}:
  {A} challenge dataset for document-level information extraction}.
\newblock In \emph{Proceedings of the 58th Annual Meeting of the Association
  for Computational Linguistics}, pages 7506--7516, Online. Association for
  Computational Linguistics.

\bibitem[{Jurgens et~al.(2018)Jurgens, Kumar, Hoover, McFarland, and
  Jurafsky}]{jurgens-etal-2018-measuring}
David Jurgens, Srijan Kumar, Raine Hoover, Dan McFarland, and Dan Jurafsky.
  2018.
\newblock \href {https://doi.org/10.1162/tacl_a_00028} {Measuring the evolution
  of a scientific field through citation frames}.
\newblock \emph{Transactions of the Association for Computational Linguistics},
  6:391--406.

\bibitem[{Kim et~al.(2010)Kim, Medelyan, Kan, and
  Baldwin}]{kim-etal-2010-semeval}
Su~Nam Kim, Olena Medelyan, Min-Yen Kan, and Timothy Baldwin. 2010.
\newblock \href {https://www.aclweb.org/anthology/S10-1004} {{S}em{E}val-2010
  task 5 : Automatic keyphrase extraction from scientific articles}.
\newblock In \emph{Proceedings of the 5th International Workshop on Semantic
  Evaluation}, pages 21--26, Uppsala, Sweden. Association for Computational
  Linguistics.

\bibitem[{Lee et~al.(2017)Lee, He, Lewis, and Zettlemoyer}]{lee2017endtoend}
Kenton Lee, Luheng He, Mike Lewis, and Luke Zettlemoyer. 2017.
\newblock \href {https://doi.org/10.18653/v1/D17-1018} {End-to-end neural
  coreference resolution}.
\newblock In \emph{Proceedings of the 2017 Conference on Empirical Methods in
  Natural Language Processing}, pages 188--197, Copenhagen, Denmark.
  Association for Computational Linguistics.

\bibitem[{Luan et~al.(2018)Luan, He, Ostendorf, and Hajishirzi}]{Luan2018}
Yi~Luan, Luheng He, Mari Ostendorf, and Hannaneh Hajishirzi. 2018.
\newblock \href {http://aclweb.org/anthology/D18-1360} {Multi-task
  identification of entities, relations, and coreference for scientific
  knowledge graph construction}.
\newblock In \emph{Proceedings of the 2018 Conference on Empirical Methods in
  Natural Language Processing}, pages 3219--3232. Association for Computational
  Linguistics.

\bibitem[{Luan et~al.(2017)Luan, Ostendorf, and
  Hajishirzi}]{luan-etal-2017-scientific}
Yi~Luan, Mari Ostendorf, and Hannaneh Hajishirzi. 2017.
\newblock \href {https://doi.org/10.18653/v1/D17-1279} {Scientific information
  extraction with semi-supervised neural tagging}.
\newblock In \emph{Proceedings of the 2017 Conference on Empirical Methods in
  Natural Language Processing}, pages 2641--2651, Copenhagen, Denmark.
  Association for Computational Linguistics.

\bibitem[{Mikolov et~al.(2013)Mikolov, Sutskever, Chen, Corrado, and
  Dean}]{word2vec_Mikolov}
Tomas Mikolov, Ilya Sutskever, Kai Chen, Greg~S Corrado, and Jeff Dean. 2013.
\newblock \href
  {https://proceedings.neurips.cc/paper/2013/file/9aa42b31882ec039965f3c4923ce901b-Paper.pdf}
  {Distributed representations of words and phrases and their
  compositionality}.
\newblock In \emph{Advances in Neural Information Processing Systems},
  volume~26. Curran Associates, Inc.

\bibitem[{Miwa and Bansal(2016)}]{miwa-bansal-2016-end}
Makoto Miwa and Mohit Bansal. 2016.
\newblock \href {https://doi.org/10.18653/v1/P16-1105} {End-to-end relation
  extraction using {LSTM}s on sequences and tree structures}.
\newblock In \emph{Proceedings of the 54th Annual Meeting of the Association
  for Computational Linguistics (Volume 1: Long Papers)}, pages 1105--1116,
  Berlin, Germany. Association for Computational Linguistics.

\bibitem[{Pezeshkpour et~al.(2020)Pezeshkpour, Tian, and
  Singh}]{pezeshkpour2020revisiting}
Pouya Pezeshkpour, Yifan Tian, and Sameer Singh. 2020.
\newblock \href {https://openreview.net/forum?id=1uufzxsxfL} {Revisiting
  evaluation of knowledge base completion models}.
\newblock In \emph{Automated Knowledge Base Construction}.

\bibitem[{Reimers and Gurevych(2019)}]{reimers-gurevych-2019-sentence}
Nils Reimers and Iryna Gurevych. 2019.
\newblock \href {https://doi.org/10.18653/v1/D19-1410} {Sentence-{BERT}:
  Sentence embeddings using {S}iamese {BERT}-networks}.
\newblock In \emph{Proceedings of the 2019 Conference on Empirical Methods in
  Natural Language Processing and the 9th International Joint Conference on
  Natural Language Processing (EMNLP-IJCNLP)}, pages 3982--3992, Hong Kong,
  China. Association for Computational Linguistics.

\bibitem[{Sun et~al.(2020)Sun, Vashishth, Sanyal, Talukdar, and
  Yang}]{sun-etal-2020-evaluation}
Zhiqing Sun, Shikhar Vashishth, Soumya Sanyal, Partha Talukdar, and Yiming
  Yang. 2020.
\newblock \href {https://doi.org/10.18653/v1/2020.acl-main.489} {A
  re-evaluation of knowledge graph completion methods}.
\newblock In \emph{Proceedings of the 58th Annual Meeting of the Association
  for Computational Linguistics}, pages 5516--5522, Online. Association for
  Computational Linguistics.

\bibitem[{Tateisi et~al.(2014)Tateisi, Shidahara, Miyao, and
  Aizawa}]{tateisi2014annotation}
Yuka Tateisi, Yo~Shidahara, Yusuke Miyao, and Akiko Aizawa. 2014.
\newblock \href
  {http://www.lrec-conf.org/proceedings/lrec2014/pdf/461_Paper.pdf} {Annotation
  of computer science papers for semantic relation extraction}.
\newblock In \emph{Proceedings of the Ninth International Conference on
  Language Resources and Evaluation ({LREC}'14)}, pages 1423--1429, Reykjavik,
  Iceland. European Language Resources Association (ELRA).

\bibitem[{Teufel et~al.(2006)Teufel, Siddharthan, and
  Tidhar}]{teufel-etal-2006-automatic}
Simone Teufel, Advaith Siddharthan, and Dan Tidhar. 2006.
\newblock \href {https://www.aclweb.org/anthology/W06-1613} {Automatic
  classification of citation function}.
\newblock In \emph{Proceedings of the 2006 Conference on Empirical Methods in
  Natural Language Processing}, pages 103--110, Sydney, Australia. Association
  for Computational Linguistics.

\bibitem[{Tsai et~al.(2013)Tsai, Kundu, and Roth}]{Tsai:2013:}
Chen-Tse Tsai, Gourab Kundu, and Dan Roth. 2013.
\newblock \href {https://doi.org/10.1145/2505515.2505613} {Concept-based
  analysis of scientific literature}.
\newblock In \emph{Proceedings of the 22nd ACM international conference on
  information \& knowledge management}, CIKM '13, pages 1733--1738, New York,
  NY, USA. ACM.

\bibitem[{Wadden et~al.(2019)Wadden, Wennberg, Luan, and
  Hajishirzi}]{wadden2019entity}
David Wadden, Ulme Wennberg, Yi~Luan, and Hannaneh Hajishirzi. 2019.
\newblock \href {https://doi.org/10.18653/v1/D19-1585} {Entity, relation, and
  event extraction with contextualized span representations}.
\newblock In \emph{Proceedings of the 2019 Conference on Empirical Methods in
  Natural Language Processing and the 9th International Joint Conference on
  Natural Language Processing (EMNLP-IJCNLP)}, pages 5784--5789, Hong Kong,
  China. Association for Computational Linguistics.

\end{thebibliography}


\end{document}